\documentclass[letterpaper, 10 pt, conference]{ieeeconf}  

\IEEEoverridecommandlockouts                         

\usepackage{microtype}
\usepackage{algorithm}
\usepackage{algorithmic}
\usepackage{amsmath,amssymb} 
\usepackage{graphicx}
\usepackage{graphics} 
\usepackage{epsfig} 
\usepackage{multirow}
\usepackage{hyperref}
\usepackage{bm}
\usepackage{caption}
\usepackage{copyrightbox}
\usepackage{subcaption}

\title{\LARGE \bf
Human-like Planning for Reaching in Cluttered Environments
}

\author{Mohamed Hasan$^{1}$, Matthew Warburton$^{2}$, Wisdom C. Agboh$^{1}$, Mehmet R. Dogar$^{1}$,\\Matteo Leonetti$^{1}$, He Wang$^{1,3}$, Faisal Mushtaq$^{2,3}$, Mark Mon-Williams$^{2,3}$ and Anthony G. Cohn$^{1}$
\thanks{$^{1}$School of Computing, University of Leeds, UK.
    }%
\thanks{$^{2}$School of Psychology, University of Leeds, UK.}%
\thanks{$^{3}$Centre for Immersive Technologies, University of Leeds, UK.}%
}

\newcommand{\figw}{1.53in} 
\newcommand{\figh}{1in} 

\begin{document}

\maketitle
\thispagestyle{empty}
\pagestyle{empty}

\begin{abstract}
Humans, in comparison to robots, are remarkably adept at reaching for objects in cluttered environments. The best existing robot planners are based on random sampling of configuration space- which becomes excessively high-dimensional with large number of objects. Consequently, most planners often fail to efficiently find object manipulation plans in such environments. We addressed this problem by identifying high-level manipulation plans in humans, and transferring these skills to robot planners. We used virtual reality to capture human participants reaching for a target object on a tabletop cluttered with obstacles. From this, we devised a qualitative representation of the task space to abstract the decision making, irrespective of  the number of obstacles. Based on this representation, human demonstrations were segmented and used to train decision classifiers. Using these classifiers, our planner produced a list of waypoints in task space. These waypoints provided a high-level plan, which could be transferred to an arbitrary robot model and used to initialise a local trajectory optimiser. We evaluated this approach through testing on unseen human VR data, a physics-based robot simulation, and a real robot (dataset and code are publicly available\footnote{\url{https://github.com/m-hasan-n/hlp.git}}). We found that the human-like planner outperformed a state-of-the-art standard trajectory optimisation algorithm, and was able to generate effective strategies for rapid planning- irrespective of the number of obstacles in the environment. 
\end{abstract}

\section{INTRODUCTION}

Imagine grasping a yoghurt tub from the back of a cluttered fridge shelf. For humans, this is a trivial task and one that even young children are able to perform exquisitely. Yet, there are a number of non-trivial questions that a robot must resolve  to execute the same action with a similar ease. Should the robot try to navigate through the available free space? Or should it start by moving obstacles? In which case, which obstacle should be moved first and to where? 

Standard robot motion planning approaches focus on identifying a collision-free trajectory that satisfies a set of given constraints \cite{latombe2012robot}, and the majority of current planning techniques are based on random sampling of the configuration space \cite{king2015nonprehensile,moll2017randomized,karaman2011sampling, elbanhawi2014sampling}. A defining feature of these sampling-based planners (SBPs) is the use of a set of probing samples drawn to uniformly cover the state space. To accelerate the planning process, it is desirable to devise non-uniform sampling strategies that favour sampling in regions where an optimal solution might lie \cite{ichter2018learning}. Finding such regions is non-trivial but, as we highlighted earlier, humans are able to find near-optimal solutions very quickly. 

Predicated on human expertise, imitation learning from demonstration (LfD) techniques are increasingly being adopted by researchers for robot motion planning \cite{fox2019multi}, \cite{zhang2018deep} \cite{rana2018towards}, \cite{tan2011computational}. For example, researchers have demonstrated the use of neural networks for learning the dynamics of arm motion from human data \cite{ravichandar2016learning}, whilst others have shown the utility of combining planning and learning-based approaches to facilitate goal-directed behaviour during human-robot interactions \cite{lawitzky2012feedback}. Alternative approaches to utilising human data include learning \textit{qualitative} task representations. Evidence indicates that humans recognise and interact with spatio-temporal space in a more qualitative than quantitative manner \cite{wallgrun2013understanding}, \cite{chen2015survey}, \cite{cohn2012thinking}. Previous work has therefore integrated qualitative spatial representations (QSRs) with manipulation planning at different levels \cite{mansouri2014more}, \cite{westphal2011guiding}. Importantly, these approaches avoid the pitfalls of SPBs which only allow a small number of objects due to the curse of dimensionality \cite{csucan2011sampling}, \cite{geraerts2004comparative}, \cite{karaman2011anytime}.

\begin{figure}
\centering 
\includegraphics[scale=0.245]{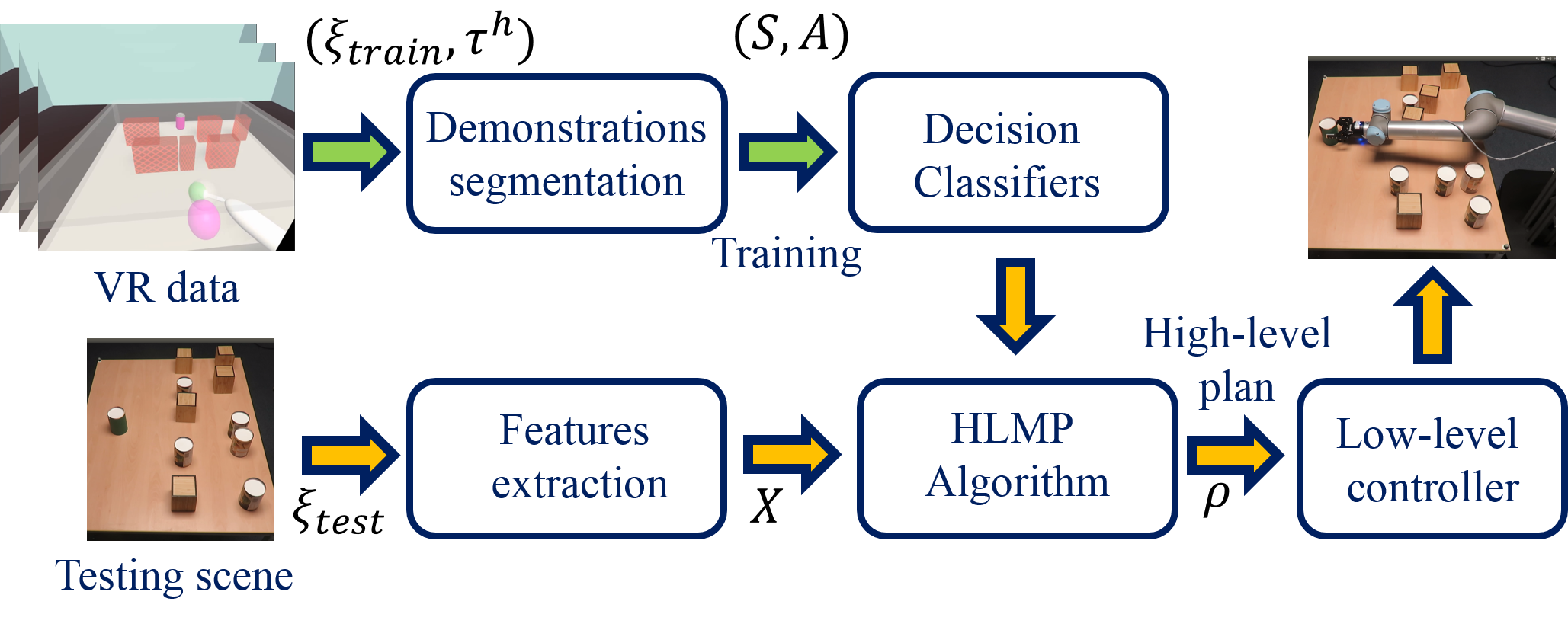}
\caption{Overview of the \emph{HLP} approach.
} 
\label{fig:overview}
\vspace{-5mm}
\end{figure}

We propose a novel approach to the problem of reaching through a cluttered environment based on geometric reasoning and workspace partitioning \cite{latombe2012robot} (similar to cell decomposition) augmented by QSRs \cite{chen2015survey}. To achieve this, we collected data from human participants for our scenario using virtual reality (\emph{VR})\cite{brookes2019studying}. Demonstrations were segmented based on a spatial model of the task space. This segmentation resulted in a set of state-action pairs used to train classifiers via standard machine learning techniques. These trained classifiers are used by a human-like planner (HLP) which, during testing, generates a high-level plan in task space. Finally, the generated high-level plan is forwarded to a robot-specific low-level controller for inverse kinematics (\emph{IK}) and local optimisation if required. The resulting method is a human-like planning (HLP) algorithm for reaching towards objects in cluttered environments (see Fig. \ref{fig:overview} for an overview of this framework).

\begin{figure*}[htb!]
\vspace{2mm}
\centering
\captionsetup{justification==centering}
\begin{subfigure}[b]{0.2\textwidth}
    \centering
    \includegraphics[height=\figh, width=\figw]{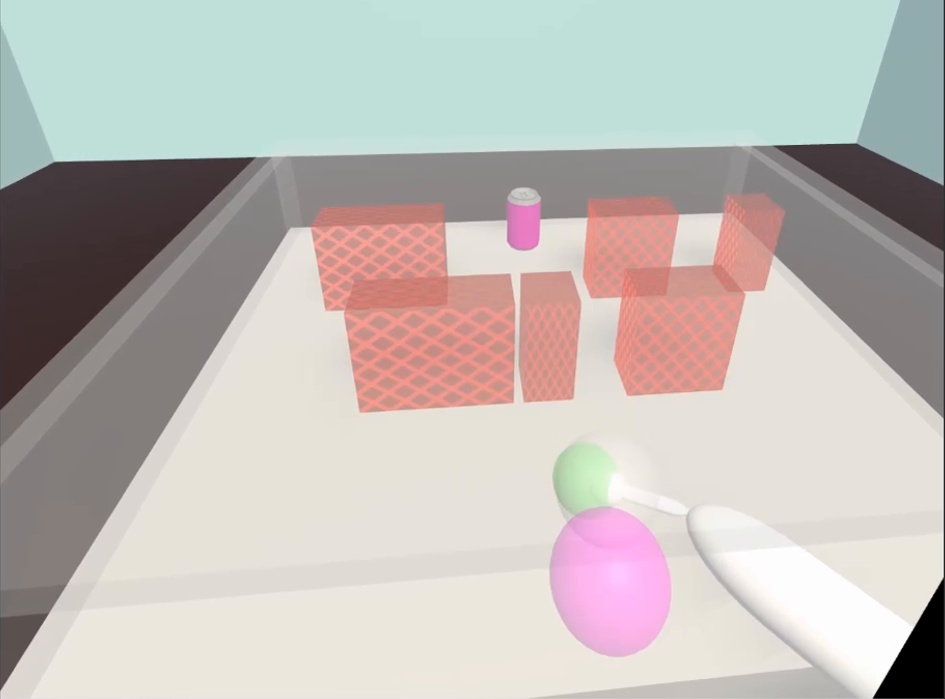}
    \caption*{Initial scene}
  \end{subfigure}
\begin{subfigure}[b]{0.2\textwidth}
  \centering
    \includegraphics[height=\figh, width=\figw]{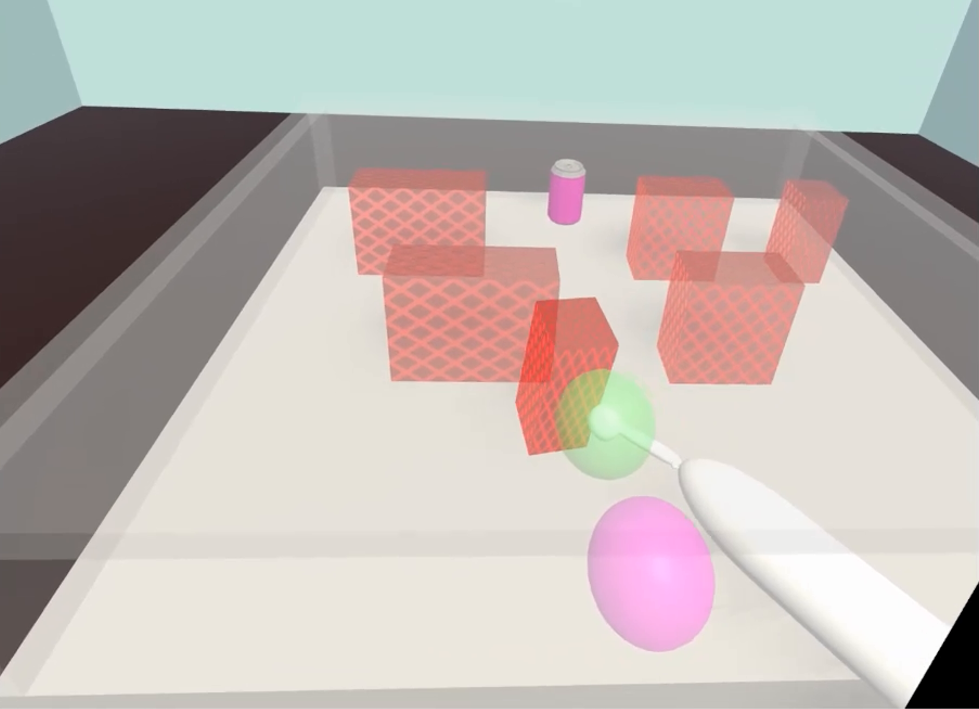}
    \caption*{Moving an object}
  \end{subfigure}
\begin{subfigure}[b]{0.2\textwidth}
\label{subfig:c}%
  \centering
    \includegraphics[height=\figh, width=\figw]{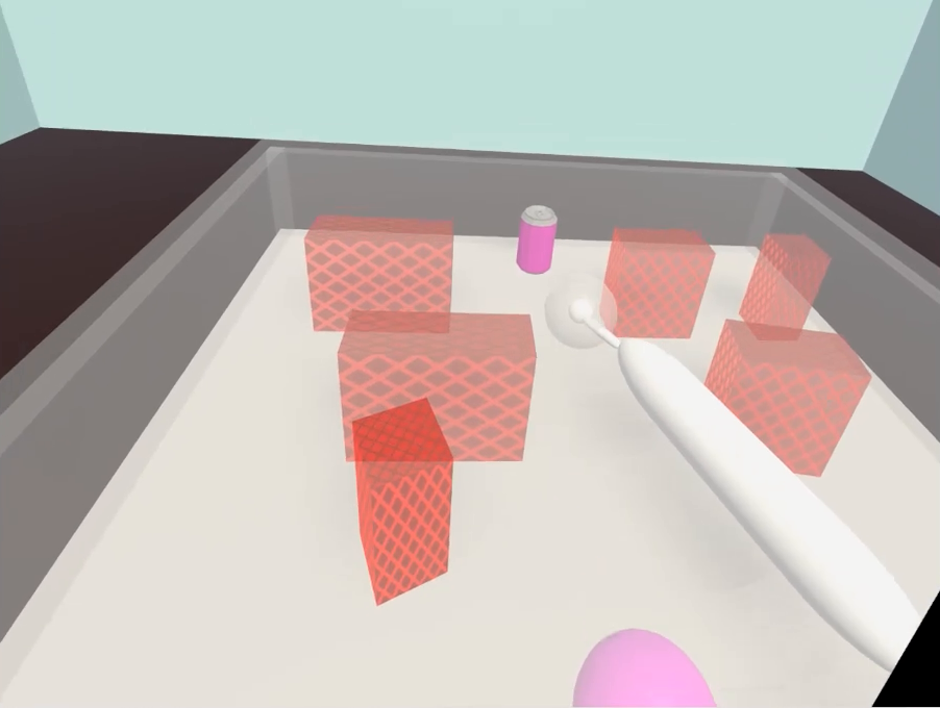}
    \caption*{Going through a gap}
  \end{subfigure} 
\begin{subfigure}[b]{0.2\textwidth}
\centering
  \includegraphics[height=\figh, width=\figw]{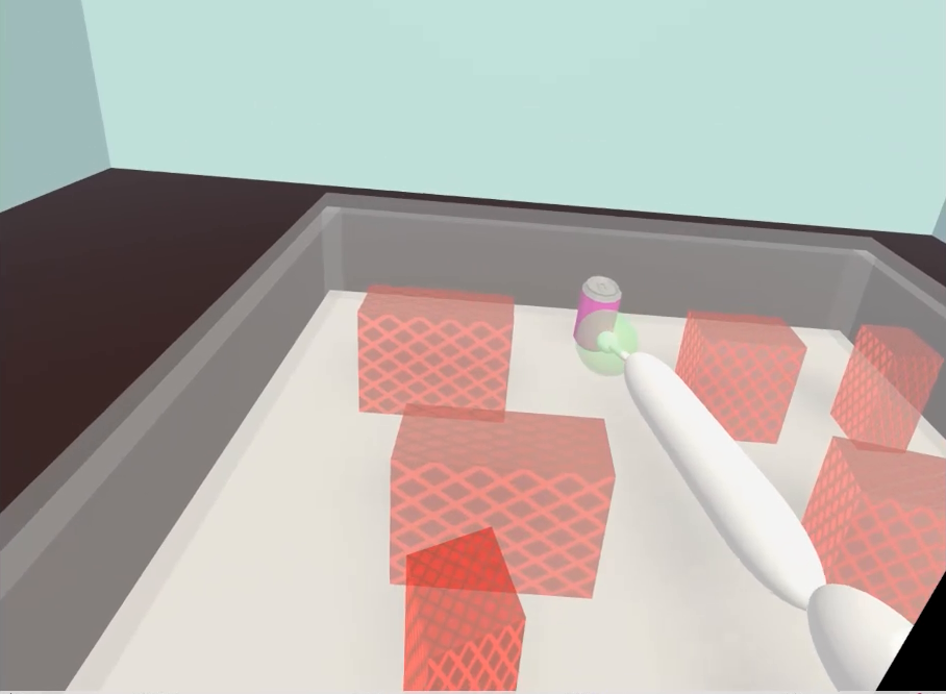}
    \caption*{Approaching the target}
  \end{subfigure}
 \caption{Sample keyframes in a VR trial. }
 \label{fig:sample_VR}
 \vspace{-4mm}
\end{figure*}

Our novel contributions include: 
\begin{itemize}
    \item The development of a new high-level planning algorithm that learns from humans operating within VR
    \item A new qualitative space/action representation to mitigate the problem of high dimensionality
    \item Empirical demonstrations of the utility of this high-level planner - showing that it is scalable and can work in conjunction with any existing low-level planner in a seamless manner.
\end{itemize}

\section{An Overview of the HLP}
Given a cluttered environment \(\xi = \{X^s, X^t, X^o_i\}\) represented by a start position \(X^s\), a target position \(X^t\) and the positions of \(N\) movable objects \(X^o_i, i=1,..., N\), we consider the problem of planning a high-level path \(\rho\) from \(X^s\) to \(X^t\) that is most likely to be selected by a human. A high-level plan \(\rho = \{X^k_i,  a^k_i\}\) is represented by a set of \(M\) keypoints \(X^k_i\) labelled by a set of associated actions \(a^k_i, i=1,..., M\).    

Our framework (Fig. \ref{fig:overview}) starts by learning from human demonstrations collected through \emph{VR}. Each demonstration is given the form of an environment representation \(\xi\) in addition to an associated human trajectory \(\tau^h\). This framework runs in two phases, training and testing. In the training phase (green arrow path of Fig. \ref{fig:overview}): given pairs of \(\{\xi_{train},\tau^h\}\), demonstrations are segmented into abstracted state-action pairs \(\{S, A\}\) that are used to train decision classifiers. In the testing phase, given \(\xi_{test}\) of a new scene, the \emph{HLP} uses the trained decision models to generate \(\rho\). The generated high-level plan can then be transferred to an arbitrary robot model. A robot-specific low-level controller solves for \emph{IK} and performs local optimisation if required.  

Demonstration segmentation is based on our modelling of the task space in order to extract required state-action pairs. Modelling of the task space is detailed in Sec. \ref{sec:model_task_space}. Feature extraction and decision classifiers are explained in Sec. \ref{subsec:classifiers} whilst the planning algorithm used in the testing phase is given in Sec. \ref{subsec:algorithm}.  Demonstration of the transfer from high-level planing to robot control is provided in the Experiments section.

\section{\emph{VR} Dataset Collection}

A dataset\footnote{\url{https://doi.org/10.5518/780}} of human demonstrations was collected by asking 24 participants to perform 90 reaching trials towards objects placed on a cluttered table top in a virtual environment (Fig. \ref{fig:sample_VR}). The table top surface was surrounded by transparent screens from all but the front side and the work space dimensions were tailored to suit adult human arm movements. The position and rotation of the hand, elbow and upper arm (nearest point to shoulder) were tracked and sampled at 90Hz.

Participants were asked to initiate each trial by first moving towards a home position, which was indicated by a transparent pink ball. After onset, a target end-point (a pink-coloured can) appeared, along with a total of six obstacles placed in two rows. Participants were instructed to interact with the scene to pick up the can and bring it back to the home position. Participants could achieve this by either navigating around the obstacles or picking up and moving the obstacles to a new position. Trials were failed and restarted if: (i) any part of the arm interacted with the obstacles; (ii) any moved object hit the edge of the workspace; (iii) the can touched any of the obstacles;  (iv) participants knocked over an obstacle with their hand.
 
\section{Modelling the Task Space}
\label{sec:model_task_space}

Devising a model of the task space that enables learning from demonstrations and generalising with different environment settings was of critical importance. To this end, we designed the row-based structure shown in Fig. \ref{fig:sample_VR} and Fig. \ref{fig:task_space}, with two rows each containing three objects and aligned start-target positions. This structured environment helped model the task space and train the decision learners. All \emph{VR} demonstrations used for training were generated according to this structure and we show in the Experiments section how the \emph{HLP} was able to generalise to different environment 
configurations.

We model the environment task space \(\xi\) qualitatively as a union of three disjoint sets: \(\xi^o\) representing the occupied space,  while the free space is given by the union of gaps \(\xi^g\) and connecting space \(\xi^c\) sets. Occupied space represents all objects in the scene in addition to walls\footnote{A set of four virtual objects are used to represent top, left, right and bottom walls and added to the occupied space to avoid objects  slipping off the surface.} of the surface on which objects rest. A gap \(g_i \in \xi^g\) is a qualitative representation of the free space existing between two objects \(o_j\) and \(o_k,  j\neq k\) on the same row.  The connecting space models the remaining free space used to connect from the start to the first row, between rows, and from the second row to the target.  

Based on this representation of the task space, the action space \(A\) is discretized into two primitive actions: \(A^o\) moving an object and \(A^g\) going through a gap (Fig. \ref{fig:sample_VR}). A primitive action occurs at a keypoint in the \(\xi^o \cup \xi^g\) space, i.e. an action applies to an object or a gap at a specific row. Actions at different rows are connected through the connecting space \(\xi^c\).     

Keypoints at objects and gaps are conceptually analogous to the idea of keyframe-based learning from demonstration \cite{akgun2012keyframe}. They are also comparable to nodes in RRTs and PRMs \cite{bry2011rapidly}, \cite{geraerts2004comparative}. Keypoints are locally connected in \(\xi^c\) in a similar manner to interpolating between keyframes or connecting graph/tree nodes.

The high-level plan is generated hierarchically at three levels: path, segment and action. At the top level, a path \(\rho\)
comprises a sequence of consecutive segments. Each segment is a qualitative description of a subset of \(\xi^c\) in which the planner (connects) interpolates between two consecutive actions. Finally, an action is one of the two primitive actions in \(A\). For example, in an environment structured in \(N_R\) rows, a path consists of \(N_R\)-\(1\) segments. Each segment connects between a pair of consecutive rows. One action takes place at a row and applies to a gap or an object. Start and target points are directly connected to the actions at the first and last rows respectively.

\section{Decision Classifiers}

\label{subsec:classifiers}
Decision learners are used to locally capture rules underlying human planning skills. We design such learners as classifiers that map from state (feature) domain \(X\) to action (class) domain \(Y\). Training examples are drawn from a distribution \(D\) over \(X \times Y\) where the goal of  each classifier \cite{moldovan2018relational}, \cite{alpaydin2009introduction}, \cite{langford2003reducing} is finding a mapping \(C:X \mapsto Y\) that minimises prediction errors under \(D\). The classifier output can be represented by the posterior probability \(P(Y|X)\). 

In this section, we introduce the features extraction required for decision classifiers. Features\footnote{Features are appropriately normalised to relevant factors such as the table size, object area, number of objects and arm dimensions.} are extracted based on the spatial representations used by humans to recognise the task space. We heavily use relational spatial representations such as distance, size, orientation and overlap. In our working scenario, there are four main decisions during a plan: (1) which gap to go through?; (2) which object to move?; (3) which direction  should the object be moved in?; (4) which segment should connect two consecutive actions?  We designed our learners as binary classifiers with the exception of 'object-moving direction'. 

\subsection{Gap and Object Classifiers (\(C_g, C_o\))}
These classifiers  have a binary output defining the probability of either selecting or rejecting a gap or an object. Intuitively, humans prefer gaps close to their hand, of an adequate size, leading to the target, and not requiring acute motions. Hence, the features vector \(X_g\) input to gap classifier \(C_g\) is defined by the distances from the gap to initial \(d_{x^g, x^s}\) and target \(d_{x^g, x^t}\) positions, gap (size) diagonal \(l_g\), orientation of the gap-to-start \(\theta_{g, s}\) and gap-to-target \(\theta_{g, t}\) lines.
\vspace{-1mm}
\begin{align}
\begin{split}
X_g = [d_{x^g, x^s}, d_{x^g, x^t}, l_g, \theta_{g, i}, \theta_{g, t}]^T 
\end{split}
\end{align}

Similarly, object features are given by distances \(d_{x^o, x^s}\) and \(d_{x^o, x^t}\), object diagonal \(l_o\), the orientation angles \(\theta_{o, s}\), \(\theta_{o, t}\) in addition to the object's overlap with the target \(l_{o,t}\) and a measure of the free space area around the object \(a_{o_{fs}}\).
\vspace{-1mm}
\begin{equation}
\begin{split}
X_o = [d_{x^o, x^s}, d_{x^o, x^t}, l_o, \theta_{o, i}, \theta_{o, t}, l_{o,t}, a_{o_{fs}}]^T \\
\end{split}
\end{equation}

\begin{figure}[t] 
\vspace{2mm}
\centering 
	\includegraphics[scale=0.4]{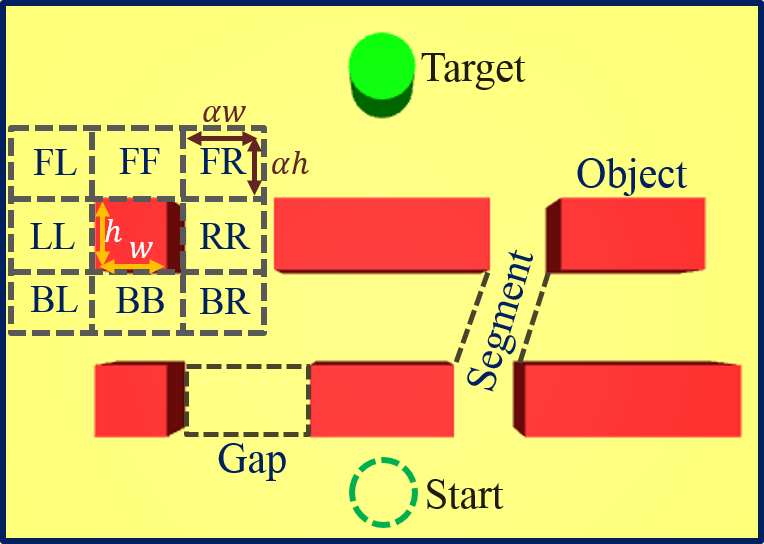} 
\caption{Modelling the task space. A 2D projection is shown for a structure defined by two rows each with three objects (hence four gaps), start location, and target object. Space around an object is discretized into eight direction classes. The size of each direction block depends on the object size and a scaling factor \(\alpha\).}
\label{fig:task_space}
\vspace{-3mm}
\end{figure}

Space around a given object is discretized into a set \(\Delta\) of eight classes (forward \emph{FF}, forward left \emph{FL}, left \emph{LL}, back left \emph{BL}, back \emph{BB}, back right \emph{BR}, right \emph{RR} and forward right \emph{FR}) covering the object's neighborhood (Fig. \ref{fig:task_space}). The size of each direction block depends on the object size and an arbitrary scaling factor. The free space in each block is computed and \(a_{o_{fs}}\) is given by the sum of free space in the eight blocks.

\subsection{Object Direction Classifier (\(C_d\))}
\label{subsub:direction_calssifier}
If the action is to move an object, we learn appropriate moving direction from the human data. To estimate the moving direction, an object is described by the direction of the human hand \(h_d \in \Delta \) when approaching the object, orientation of an object-to-target line \(\theta_{o,t}\) and the amount of free space \(a_{fs}\) in each surrounding direction around the object. This is a multi-class classifier whose output \(Y_d = \Delta\). 
\vspace{-1mm}
\begin{equation}
\begin{split}
X_d = [h_d, \theta_{o,t}, a_{fs} ]^T \\
\end{split}
\end{equation}

\vspace{-3mm}
\subsection{Connecting Segments Classifier (\(C_c\))}
\label{subsub:segment_calssifier}
Each segment connects two actions applied on gaps and/or objects. Let \(e_1\) and \(e_2\) denote the elements representing these gaps/objects at the first and second rows respectively. It is desirable to avoid actions that are considerably apart from each other, having no or small overlap with the target and expected to result in a large collision. Hence, a segment feature vector consists of the signed horizontal \(d_{{(e1, e2)}_x}\) and vertical \(d_{{(e1, e2)}_y}\) distances, overlap \(l_{c, t}\) between \(d_{{(e1, e2)}_x}\) and the target object, segment orientation \(\theta_c\) w.r.t a straight line connecting initial and target positions, and the collision \(c_\zeta\) expected to accompany motion through this segment. 

Segment collision is computed as the overall overlapping volume (area) between human forearm (link) and \(\xi^o\). The overlapping area is approximated by the intersection polygon area and found by scanning the surrounding area using sampling lines.\footnote{Although there are standard graphics polygon clipping algorithms to solve this kind of problem, we preferred an approximate but faster line-sampling approach.} 
\vspace{-1mm}
\begin{equation}
\begin{split}
X_c = [d_{{(e1, e2)}_x}, d_{{(e1, e2)}_y}, l_{c, t}, \theta_c c_s]^T 
\end{split}
\end{equation}

\section{\emph{HLP} ALGORITHM}
\label{subsec:algorithm}

The \emph{HLP} algorithm uses the trained classifier models explained in the previous section to generate the high-level plan in the testing phase. The algorithm starts by locating rows \(R_i , i=1, ..., N_R\) and gaps \(\xi^g\) in the scene. 

For each \(i\)-th row, object and  gap classifiers are called (Lines 2-3) to identify the best \(N_{i_o}\) objects and \(N_{i_g}\) gaps respectively. Selected gaps and objects in the (\(i\))-th row are connected to their counterparts in the next (\(i+1\))-th row through connecting segments. Selecting a total \(N_g = \sum_{j=i}^{i+1} N_{j_g}\) gaps and \(N_o = \sum_{j=i}^{i+1} N_{j_o}\) objects in a pair of consecutive rows results in \(N_g N_o\) segment combinations. These combinations are constructed and passed to the segment classifier that selects the best \(N_c\) connecting segments (Lines 4-6).   

\begin{algorithm}[H]
 \caption{The Human-Like Planner (HLP)}
 \begin{algorithmic}[1]
 
 \REQUIRE Environment representation \(\xi = \{X^s, X^t, X^o_i\}\)
 
 \ENSURE  High-level path \(\rho\)\\
 
 Locate rows $R$ and gaps \(\xi^g\)
 \FORALL{$R$} 
 \STATE
 {
 Compute gaps feature vector $X_g$
 
 $G_{selected}\gets {C_g(X_g)}$
 
 Compute objects feature vector $X_o$
 
 $O_{selected}\gets {C_o(X^o_i)}$
 } 
 \ENDFOR
 
 \FORALL{pairs of consecutive rows} 
 \STATE
 {
 $C\gets$ Segment Constructor ${(G_{selected}, O_{selected})}$
 
 Compute segments feature vector $X_c$
 
 $C_{selected}\gets {C_c(X_c)}$
 } 
 \ENDFOR
\FORALL  {$C_{selected}$} 
 \STATE{   \IF {$a^o\in C_{selected}$}
       \STATE {Compute object-direction feature vector $X_d$\\
       Object direction = ${C_d(X_d)}$}\\
       Augment $C_{selected}$ by expected object's location
 \ENDIF\\
 Compute arm configuration feature vector $X_a$\\
 Estimate arm configuration: ${R_a(X_a)}$\\
 Compute expected path collision \(\rho_\zeta\)}
 
 \ENDFOR
 
 Select the path with minimum collision score
 
 \label{alg:HLMP}
 
 \end{algorithmic}
\end{algorithm}
\vspace{-2mm}

\begin{figure*}[htb!]
\vspace{2mm}
\centering
\captionsetup{justification==centering}
\begin{subfigure}[b]{0.18\textwidth}
    \centering
    \includegraphics[height=\figh, width=\figw]{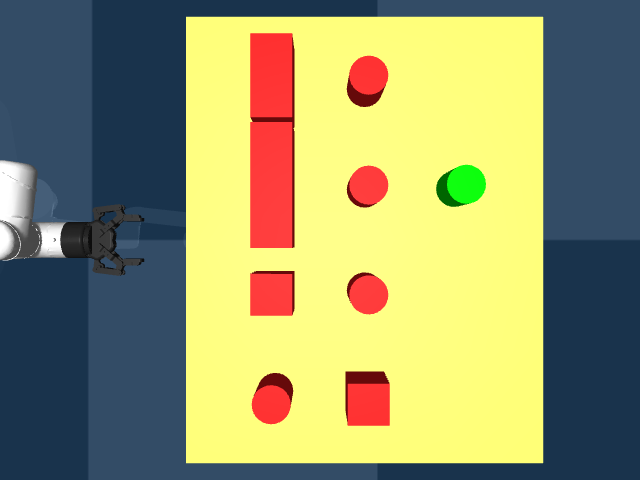}
  \end{subfigure}
\begin{subfigure}[b]{0.18\textwidth}
  \centering
    \includegraphics[height=\figh, width=\figw]{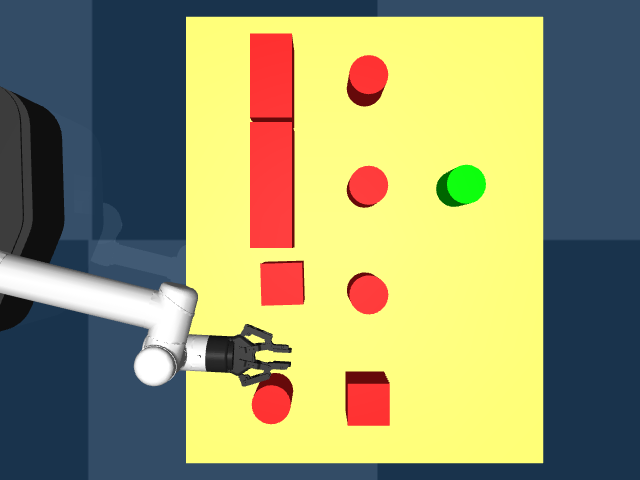}
  \end{subfigure}
\begin{subfigure}[b]{0.18\textwidth}
  \centering
    \includegraphics[height=\figh, width=\figw]{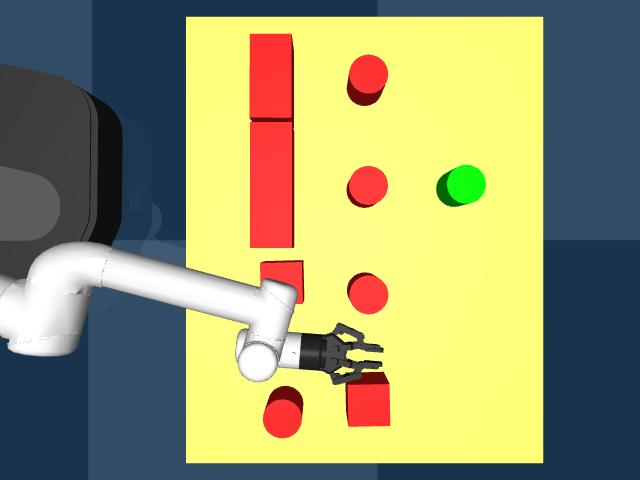}
  \end{subfigure} 
\begin{subfigure}[b]{0.18\textwidth}
\centering
   \includegraphics[height=\figh, width=\figw]{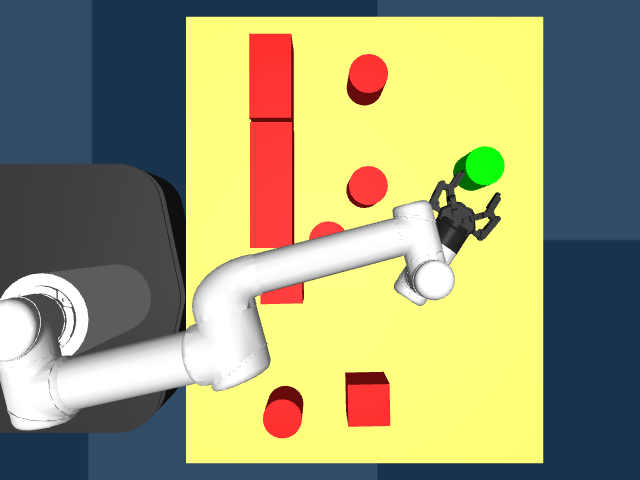}
  \end{subfigure}
 \caption{A human-like manipulation plan for reaching a target amongst 8 obstacles. HLP is shown to generalize to different configurations in terms of  table size, number and shape of objects.}
 \label{fig:sample_sim_execution}
 \vspace{-4mm}
\end{figure*}


A segment connects two actions, one of which may belong to the moving-object class. Hence, the object moving direction is estimated by the \(C_d\) classifier using the same convention of space discretization in Sec. \ref{subsub:direction_calssifier}. The expected object location after moving is found and added to the segment's sequence of keypoints (Lines 7-11).  

Next, the human-arm configuration is estimated (see next section) at each keypoint, and estimated configurations are used to evaluate the overall expected collision of each segment (Lines 11-12). To this end, we get candidate segments between rows that are labelled by a measure of their overall expected collision. For a two-row structure, the best segment is picked as the one with least likely collision.
 
\subsection{Arm Configuration and Collision Detection}
In the 2-row structure, a candidate path has one segment that is connected to the start and target points. Having a number of candidate paths, the algorithm  selects the one with minimum overall expected collision. Overall expected path collision is found by computing collision of full-arm motion with \(\xi^o\) between path keypoints. Therefore, arm configurations are firstly estimated at each keypoint and then expected arm collision is computed. 

A human arm is modeled as a planar arm with four joints at neck, shoulder, elbow and hand. The arm configuration is represented by two angles: \(\theta_{sh}\) between neck-shoulder and upper arm links, and \(\theta_{el}\) between upper arm and forearm links. The arm configuration at a given keypoint \(K_t\) is estimated by regression. Input features to the configuration regression model \(R_a\) are: hand direction \(h_d \in \Delta \) when approaching \(K_t\), arm configuration at the previous keypoint \(K_{t-1}\) and signed horizontal and vertical distances between the two keypoints. 
\begin{equation}
\begin{split}
X_a = [h_d, \theta_{{sh}_{t-1}}, \theta_{{el}_{t-1}}, d_{{(k1, k2)}_x},  d_{{(k1, k2)}_y}]^T \\
\end{split}
\end{equation}
\vspace{-3mm}

Through estimating arm configurations at keypoints (and hence joint positions), full arm collision is computed as the sum of its link collisions during motion along keypoints from the start to target. The collision of each link is computed using the same approach as in Sec. \ref{subsub:segment_calssifier}. Intersections between sampling lines and objects are found and the area of the resulting intersection polygon defines the collision amount. To this end, a number of candidate paths are generated, each labelled with a measure of its overall expected collision \(\rho_\zeta\). This step completes the \emph{HLP} algorithm and generates a plan defined by the path having minimum expected overall collision.  

\vspace{1mm}
\section{EXPERIMENTS AND RESULTS}
The \emph{HLP} approach was tested through three protocols: human \emph{VR} data, robot simulation and real-robot experiments. The \emph{VR} dataset was randomly split into two disjoint sets: approximately 80\% (19 participants and 1710 trials) for training, and 20\% (5 participants and 450 trials) for testing.  This \textit{cross-participant splitting}, (i.e. no overlap between training and testing data in terms of participants) allowed us to test the generalisation of the proposed approach with participants who had not been seen in training. This splitting was repeated for five folds of randomly selected participants for both sets. Standard (MATLAB) support vector machines with Gaussian kernel and Gaussian process regression were used to implement classifiers and regression models respectively. The same trained models and planning algorithm were used for all experiment protocols. \vspace{-1mm}     

\begin{center}
\begin{table}[b]
\caption{Results (mean and standard deviation) of the 5-fold \emph{VR} test experiment.} 
\centering 
\begin{tabular}{c c c} 
\hline\hline 
Metric & Mean & STD\\ 
\hline \hline 
\(C_g\) accuracy  & 0.95  & 0.002\\
\(C_o\) accuracy  & 0.85 & 0.005 \\
\(s_{HLP}\) (overall) & 0.70 & 0.011 \\ 
\(s_{HLP}\) (\(I(D_n)\)) & 0.79 & 0.016 \\
\(s_{HLP}\) (\(I(E_n)\)) & 0.67 & 0.012\\
\hline\hline
\end{tabular}
\label{tab:vr_test}
\end{table}
\vspace{-5mm}
\end{center}

\subsection{Experiments on \emph{VR} Data}
This protocol tested the \emph{HLP} on unseen, but similarly distributed data to the training set. The objective was to measure the similarity of the generated plans to the \textquotedblleft ground truth\textquotedblright \space human plans. This similarity was computed for a scene having \(N_R\) rows by the following metric:
\begin{align}
\begin{split}
s_{HLP} = \frac{1}{2N_R} \sum_{n=1}^{N_R} I(D_n)(I(D_n)+I(E_n))
\end{split}
\end{align}
where \(I(.)\) is the indicator function which is \(1\) when its argument is true and \(0\) otherwise, \(D_n\) is a Boolean variable that is true if \emph{HLP} action is same as human action at the \(n\)-th row, and \(E_n\) is a Boolean variable that is true if the same element (gap or object) is selected by both \emph{HLP} and human action. To illustrate, if both the \emph{HLP} and the human participant decided to move an obstacle in the first row and then go through a gap in the second row, then this would be quantified as a \(50\%\) similarity rate. This could increase to \(100\%\) if both selected the same obstacle and the same gap. 

Mean and standard deviation of the 5-fold results are shown in Table \ref{tab:vr_test}. Accuracy of the gap and object classifiers are 95\% and 85\% respectively. It is worth noting that we compared similarity of the \emph{HLP} output to a specific participant's plan at a time and then reported the mean similarity. This means that \emph{HLP} similarity is counted if it exactly matches a specific testing participant who was never seen during training. On average, our planner was 70\% similar to the test participant plans. More specifically, the \emph{HLP} selected the same action as a human 79\% of the time, while it selected the same element (specific gap or object) 67\% of the time.

\subsection{Robot Experiments}
Through robot experiments, we compared the \emph{HLP} with a standard physics-based stochastic trajectory optimisation (STO) approach \cite{Agboh_Dogar_WAFR_2018, Agboh_Dogar_Humanoids_2018, Agboh_Daniel_Dogar_ISRR_2019}. These algorithms were implemented using the Mujoco \cite{mujoco} physics engine and the Deepmind control suite \cite{dm_control}. We assumed a world consisting of objects on a table with a 10-DOF robot as shown in Fig. \ref{fig:sample_sim_execution}. As a baseline algorithm, we used a state-of-the-art physics-based stochastic trajectory optimiser \cite{Agboh_Dogar_WAFR_2018}, initialised with a straight line control sequence from the end-effector's start pose to the target object. 

\emph{IK} solutions for the high-level plan keypoints were found and connected with straight lines in the robot's configuration space to generate a control sequence. This was passed to a trajectory optimiser as an initial candidate control sequence. Thus, for the \emph{HLP}, the number of actions in a given control sequence varied depending on the high level plan. In contrast, the baseline approach (STO) initialised the trajectory optimiser using a straight line control sequence to the goal. 

We compared performance of the \emph{HLP} and \emph{STO} quantitatively through success rates and planning times of simulation experiments, and qualitatively through visual inspection of manipulation plans in real robot experiments. 

\subsubsection{Robot Simulation Experiments}
Simulation experiments evaluated the performance of our planner in scenes generated from a distribution considerably different to that of the training data. Generalisation was investigated across different numbers, sizes and types of objects. A plan was classified as being successful if the target object was inside the gripper's pre-grasp region at the final state, and if no other object dropped off the table during the trial. We recorded the total planning time for successful trials of each planner. Planning time comprised an initial time used to generate the high level plan (straight line for $STO$) and an optimisation time. Two simulation categories were considered:

\begin{figure*}[htb!]
\vspace{2mm}
\centering
\captionsetup{justification==centering}
\hspace{10mm}
\begin{subfigure}[b]{0.17\textwidth}
    \centering
    \begin{picture}(0,0)
          \put(-60,-45){\rotatebox{90}{HLP(1X)}}
    \end{picture}
    \includegraphics[scale=0.08]{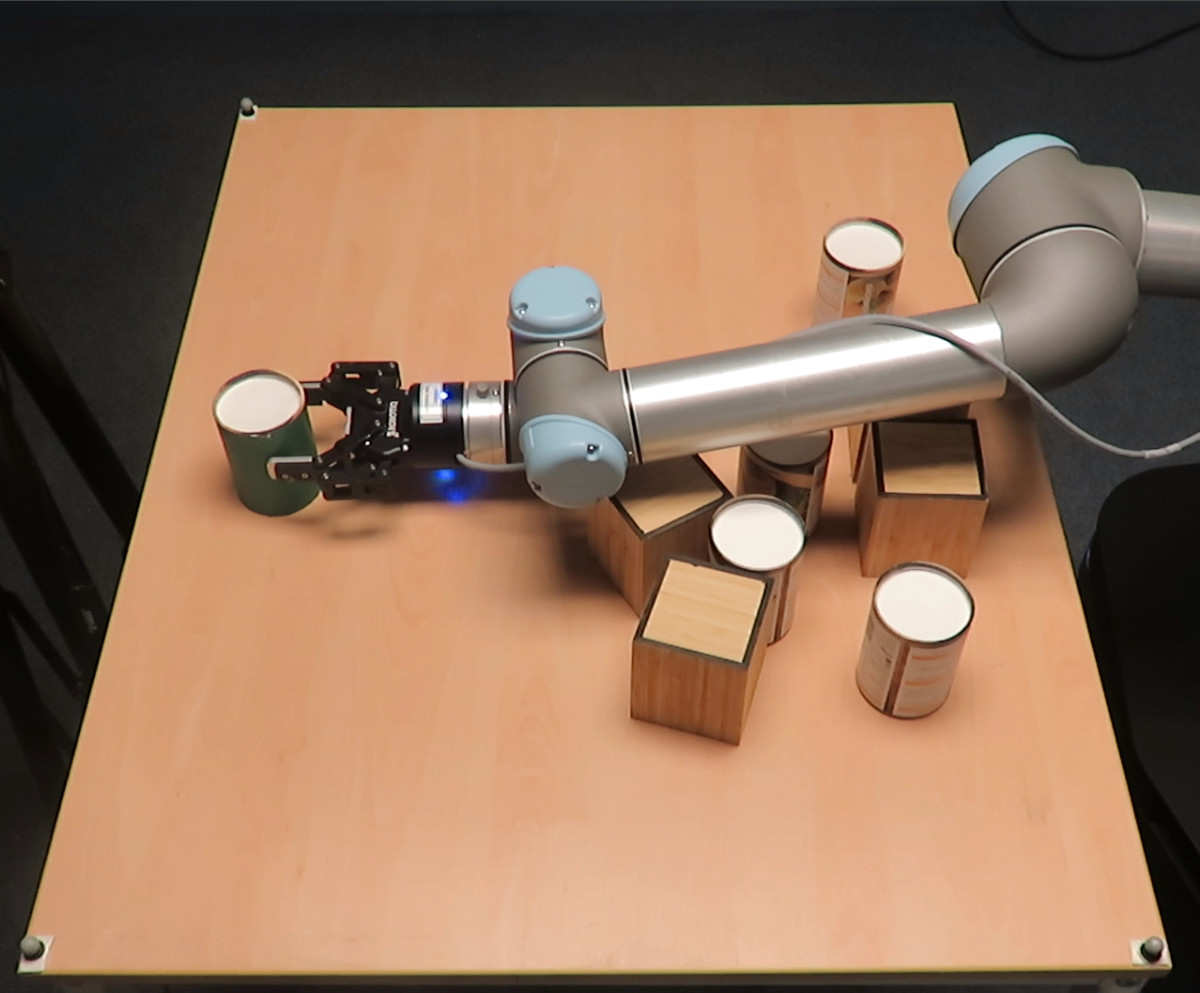}
  \end{subfigure}
\hspace{-3mm}
\begin{subfigure}[b]{0.17\textwidth}
  \centering
    \includegraphics[scale=0.08]{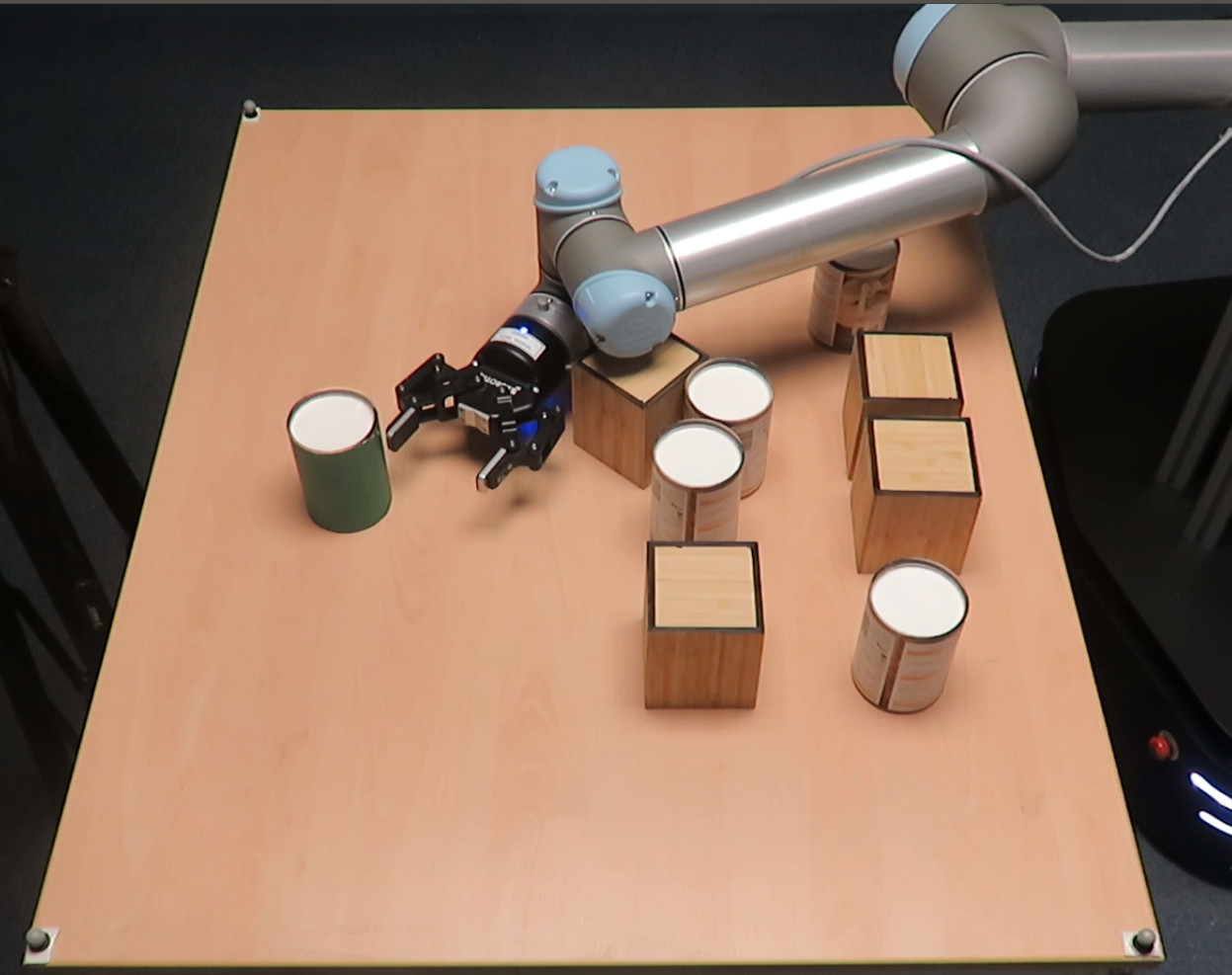}
  \end{subfigure}
\hspace{-3mm}
\begin{subfigure}[b]{0.17\textwidth}
  \centering
    \includegraphics[scale=0.08]{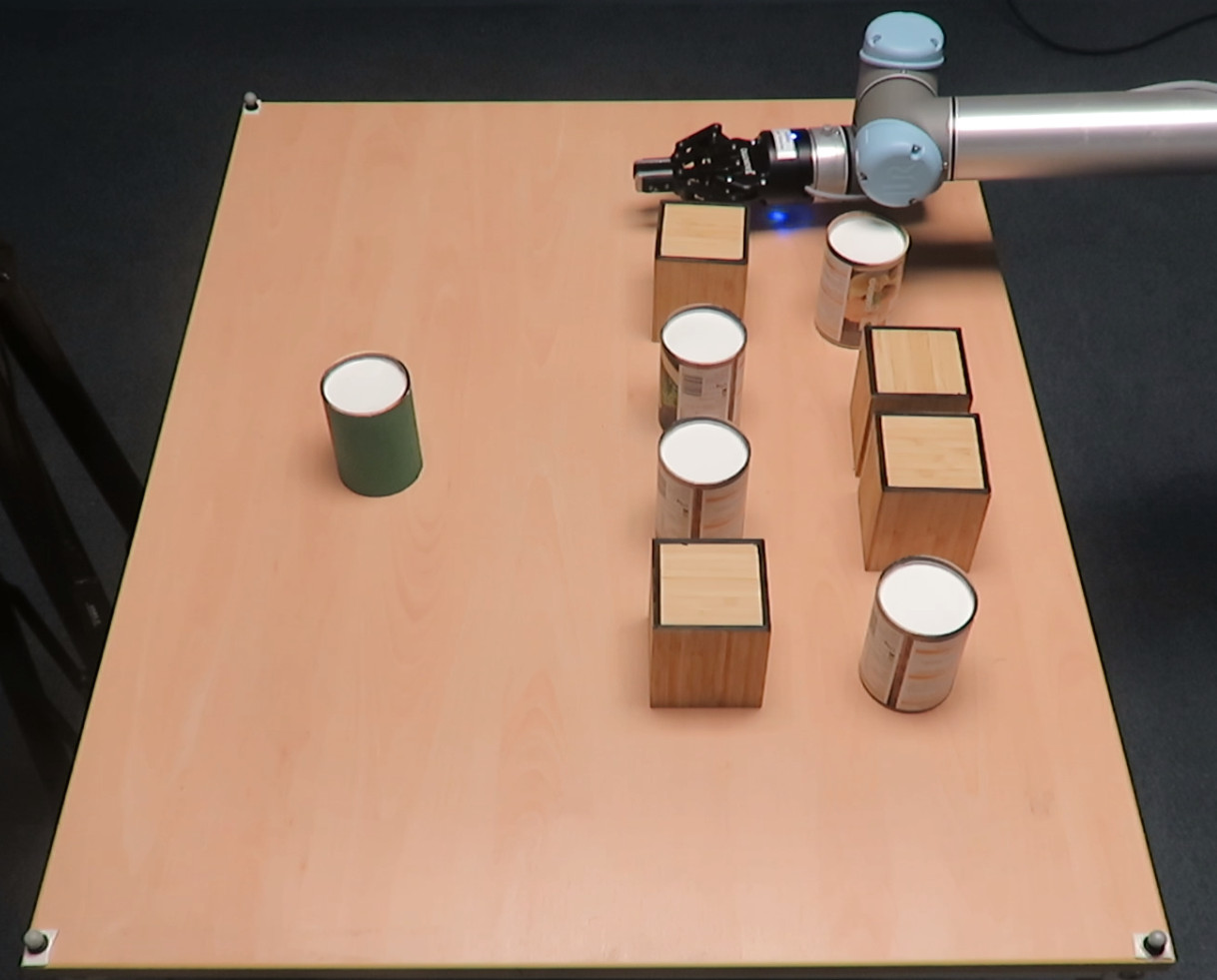}
  \end{subfigure} 
\hspace{-3mm}
\begin{subfigure}[b]{0.17\textwidth}
\centering
   \includegraphics[scale=0.08]{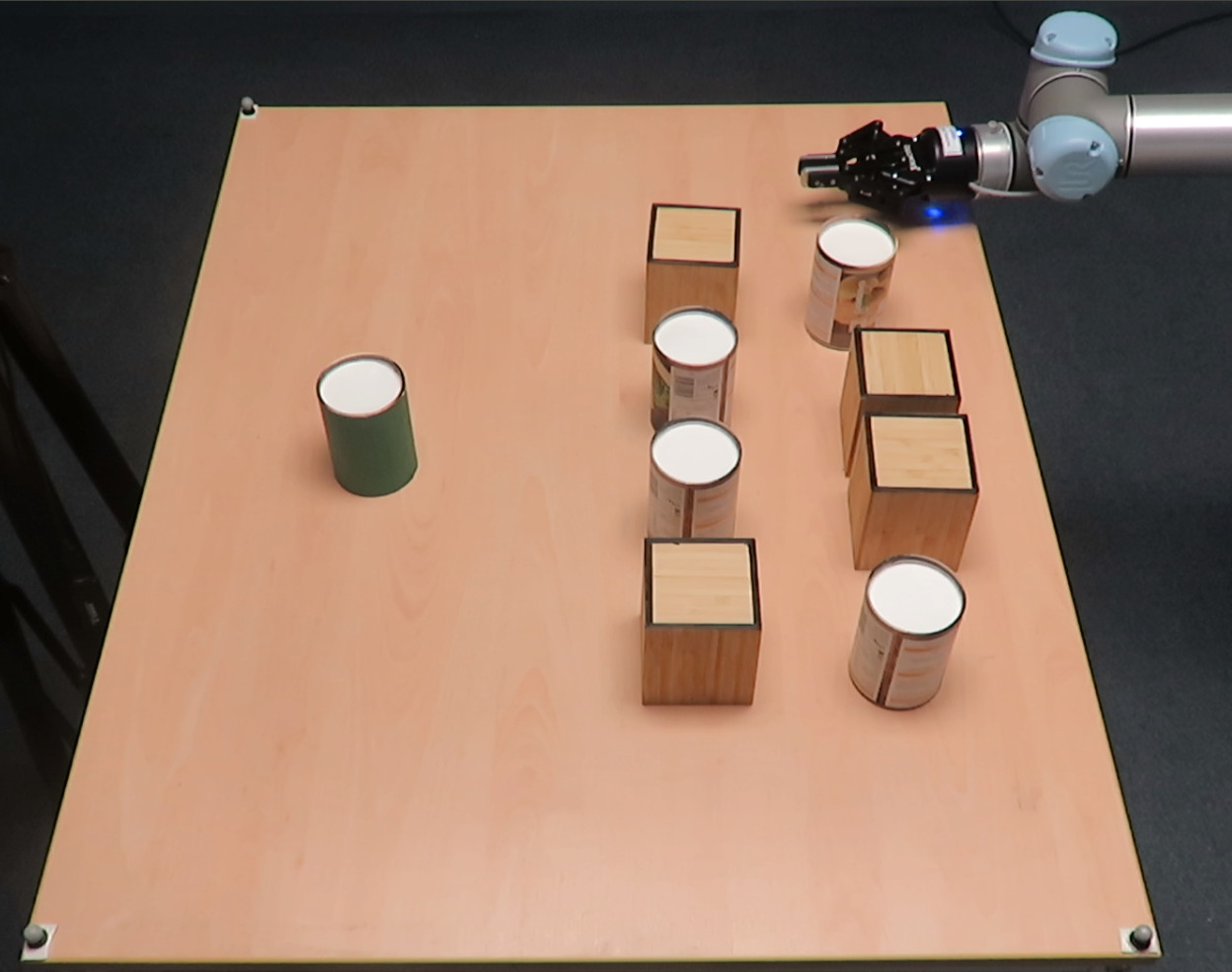}
  \end{subfigure}
 \hspace{-3mm}
  \begin{subfigure}[b]{0.17\textwidth}
\centering
   \includegraphics[scale=0.08]{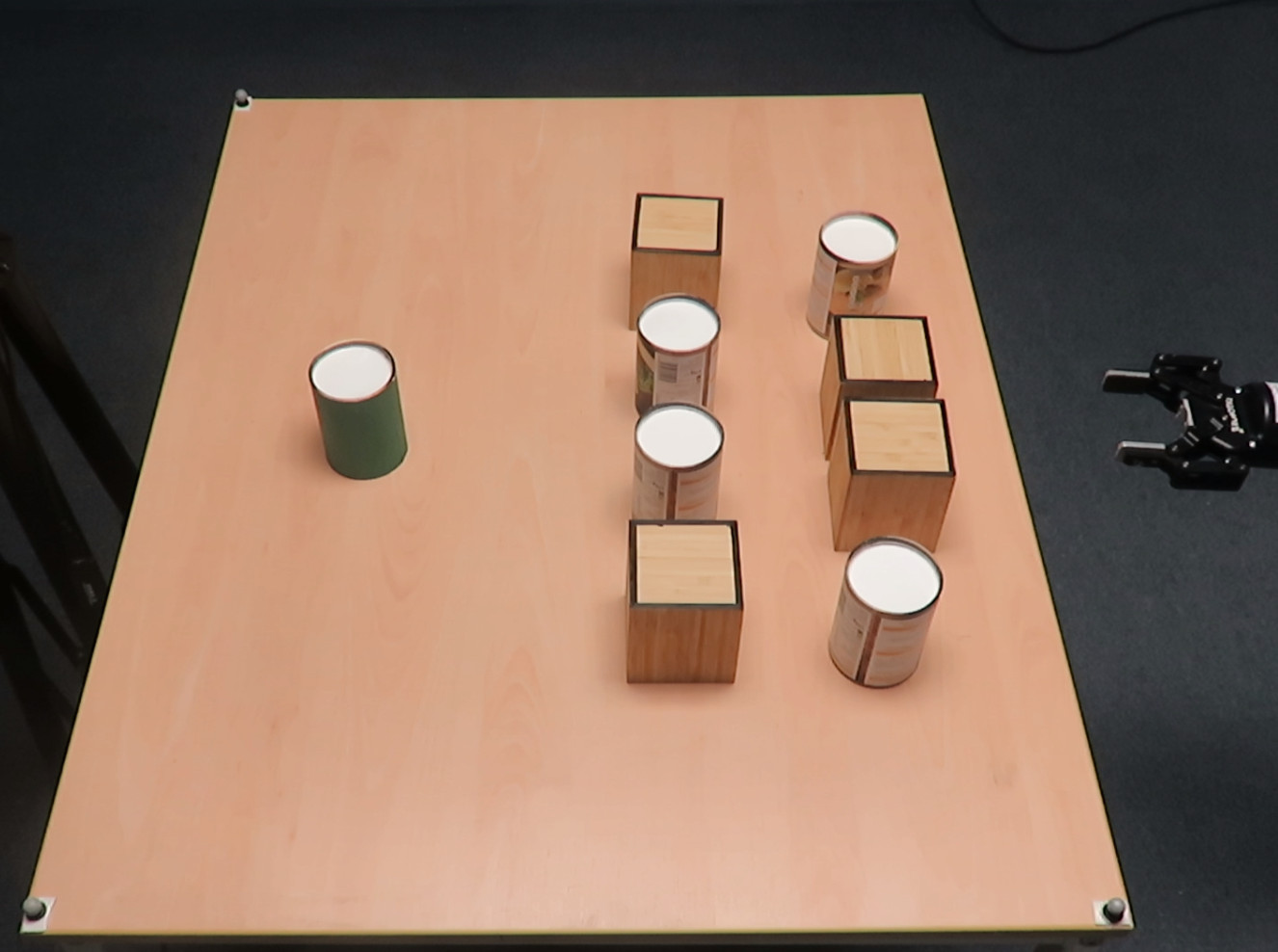}
  \end{subfigure}

  \hspace{10mm}
  \begin{subfigure}[b]{0.17\textwidth}
    \centering
        \begin{picture}(0,0)
          \put(-60,-45){\rotatebox{90}{STO(1.5X)}}
    \end{picture}
    \includegraphics[scale=0.08]{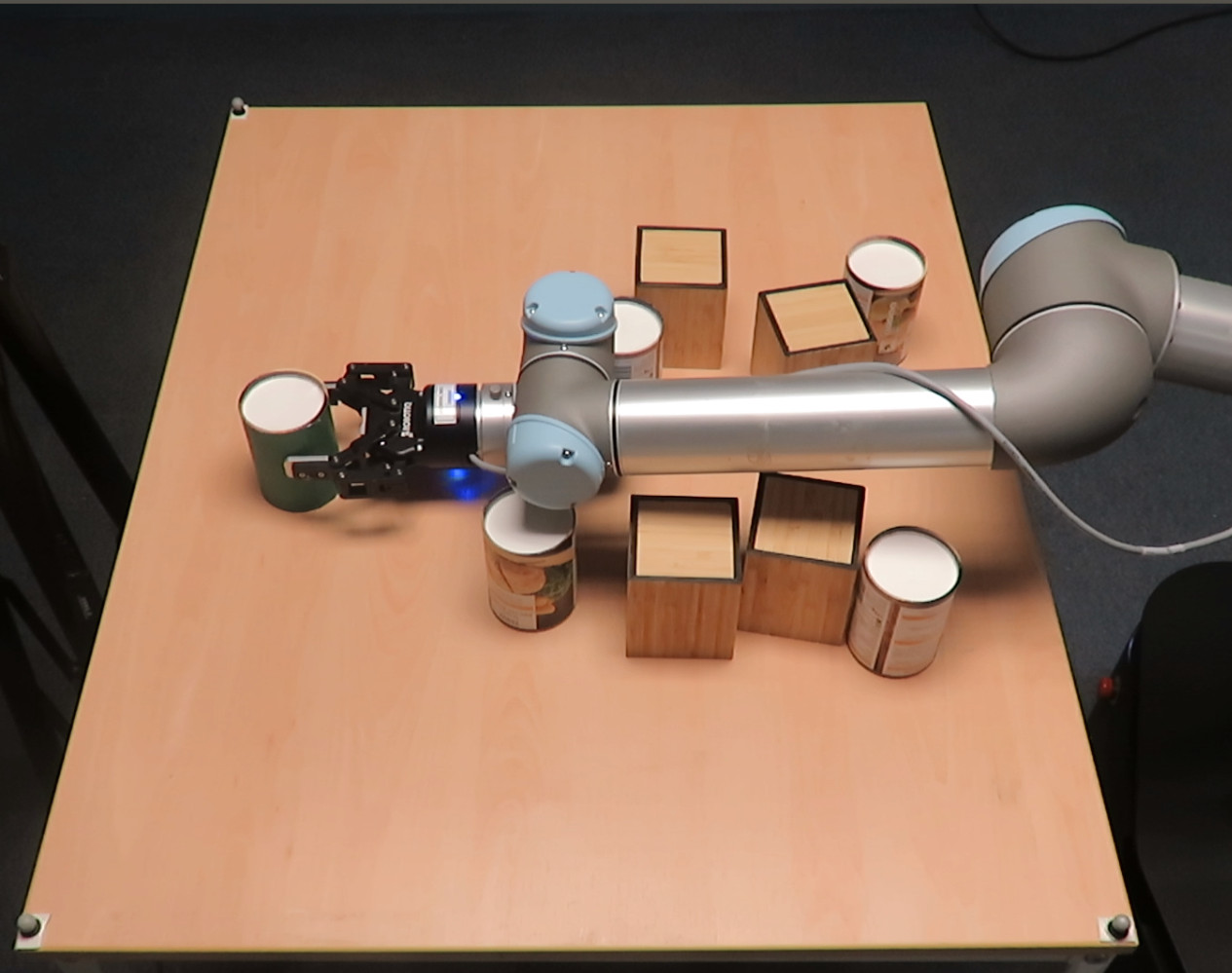}
  \end{subfigure}
 \hspace{-3mm}
\begin{subfigure}[b]{0.17\textwidth}
  \centering
    \includegraphics[scale=0.08]{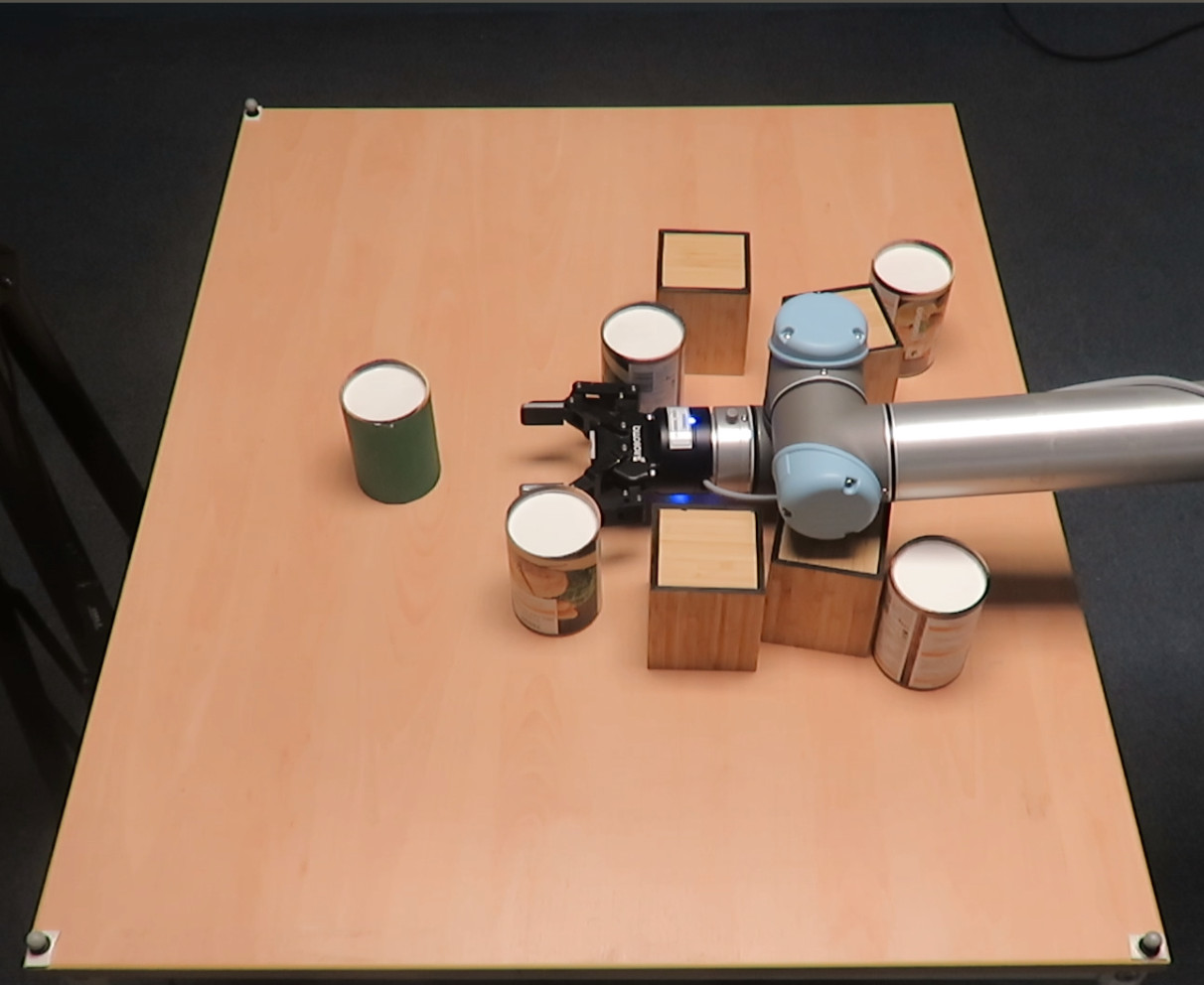}
  \end{subfigure}
\hspace{-3mm}
\begin{subfigure}[b]{0.17\textwidth}
  \centering
    \includegraphics[scale=0.08]{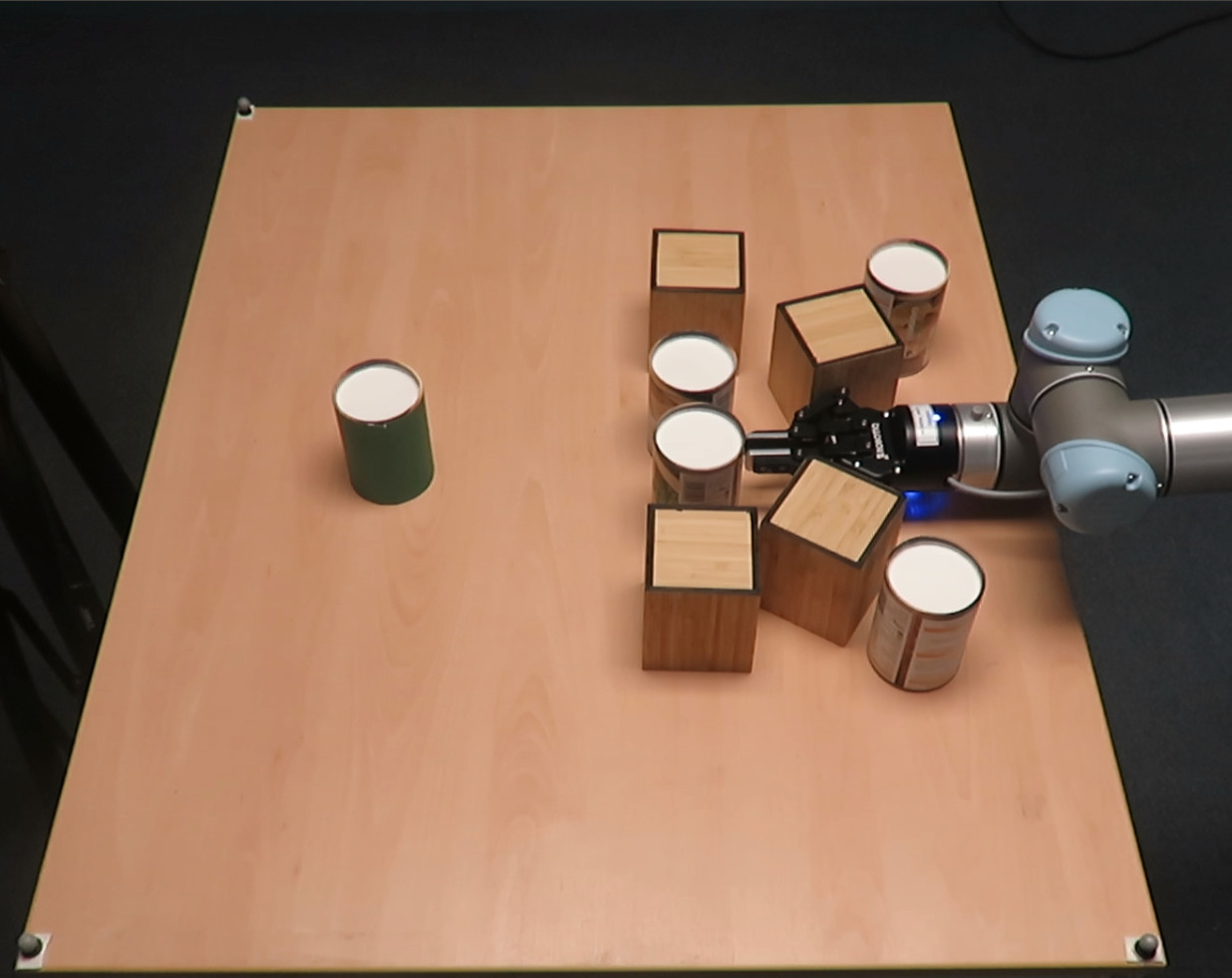}
  \end{subfigure} 
\hspace{-3mm}
\begin{subfigure}[b]{0.17\textwidth}
\centering
   \includegraphics[scale=0.08]{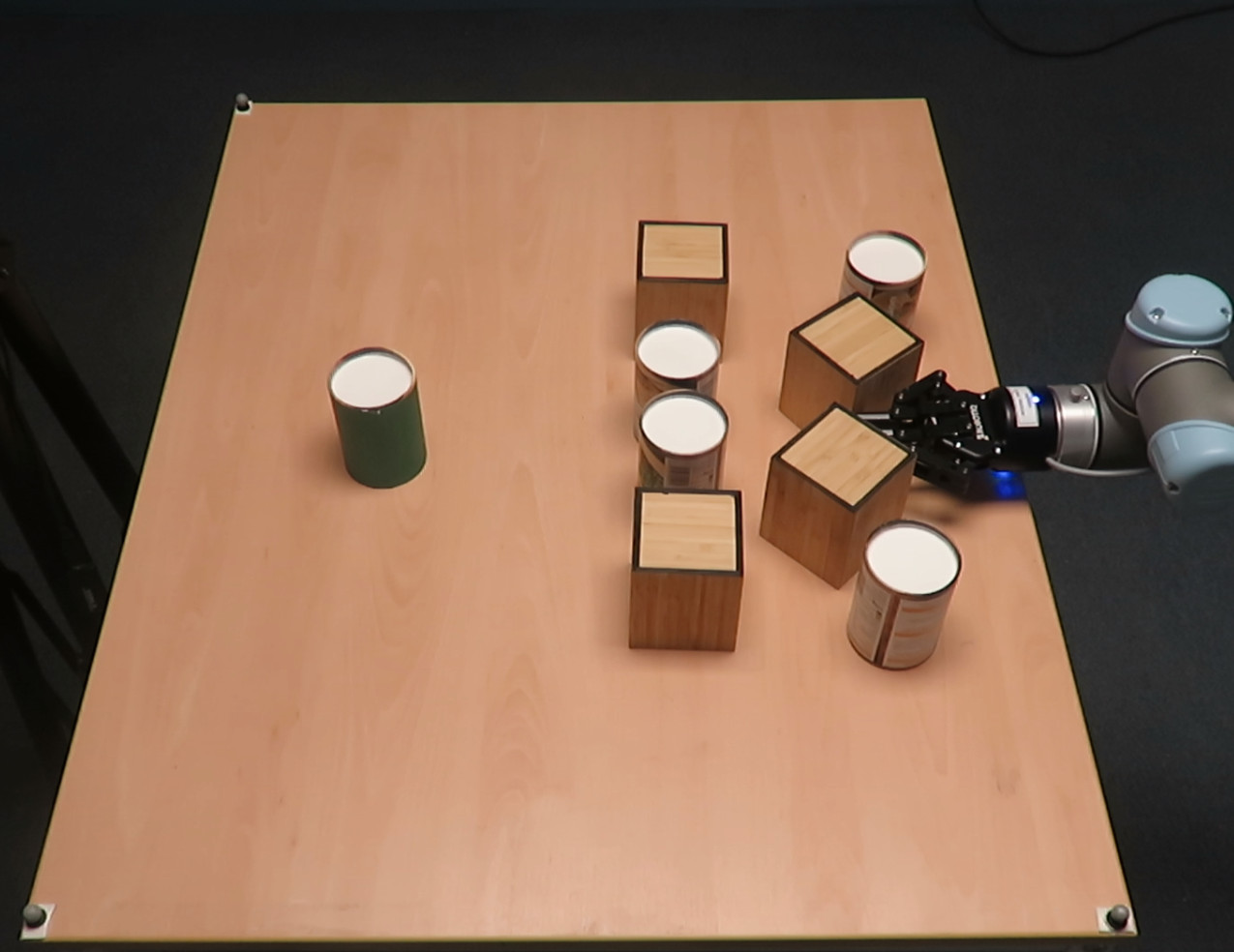}
  \end{subfigure}
  \hspace{-3mm}
  \begin{subfigure}[b]{0.17\textwidth}
\centering
   \includegraphics[scale=0.08]{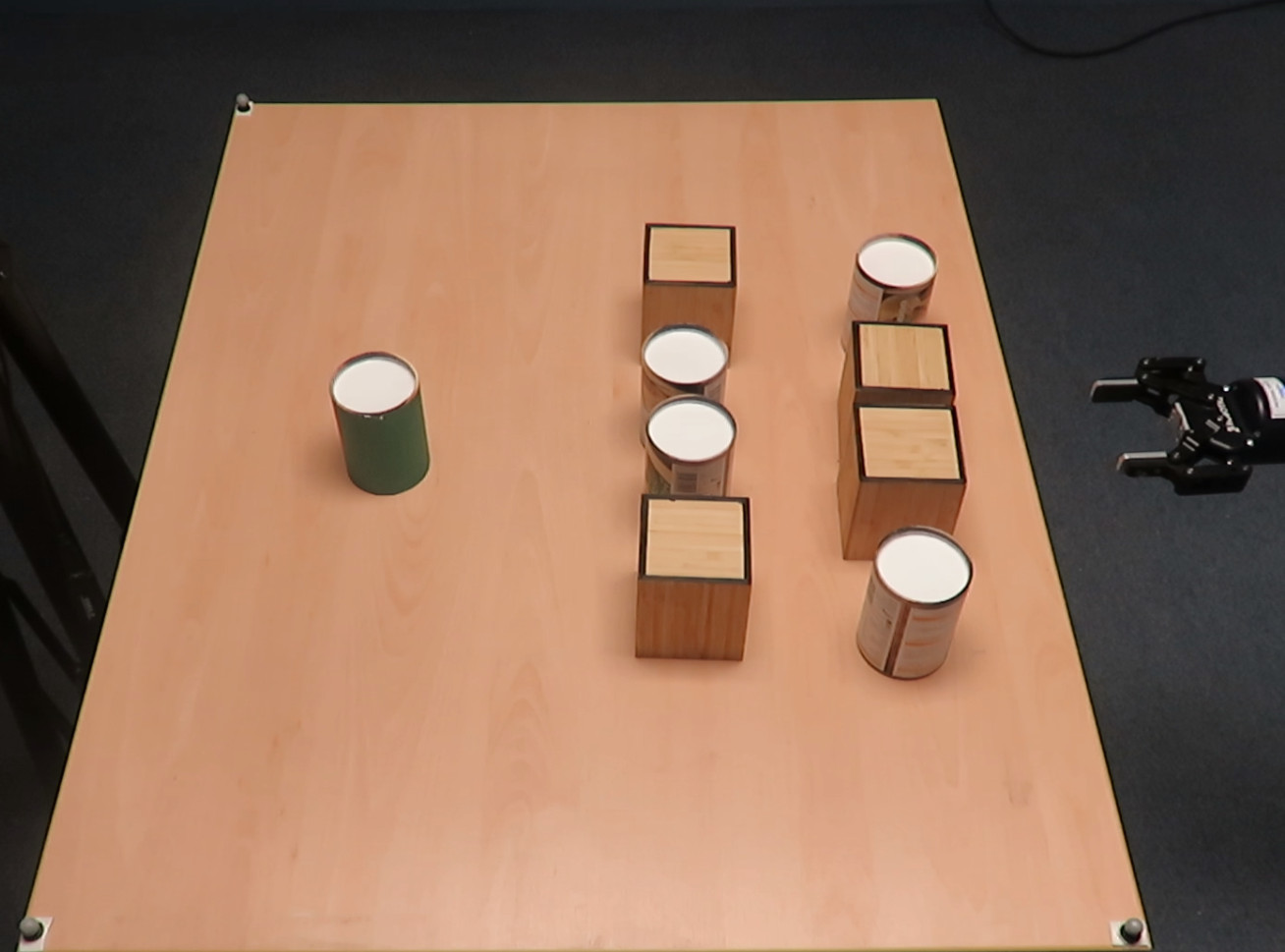}
  \end{subfigure}
 \caption{A human-like manipulation plan (Top; HLP), and a baseline plan (Bottom; STO). HLP can reason over task space and navigate through the available free space. On the other hand, STO is biased towards its straight-line initialization and thus had to push objects around.}
 \label{fig:real_robot}
\end{figure*}
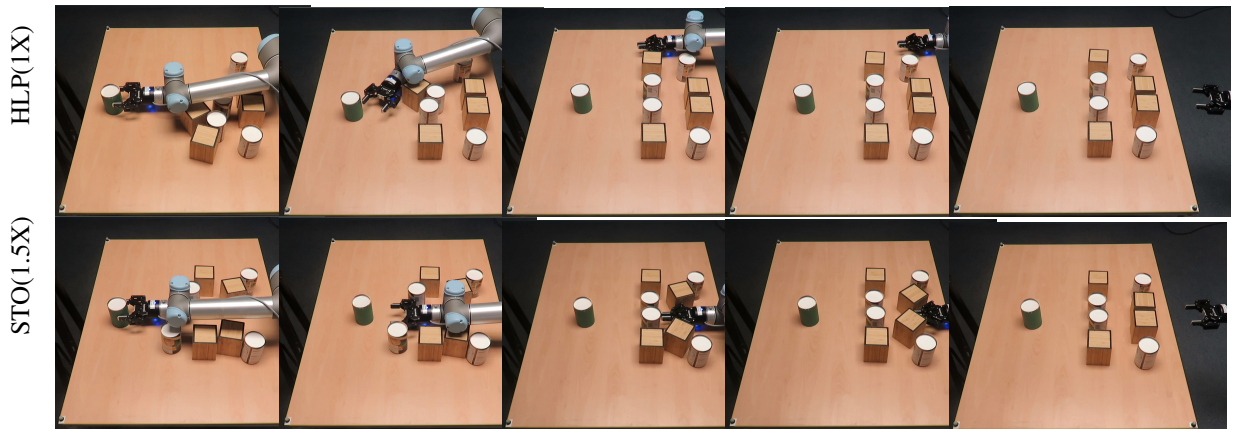

\textbf{\emph{Generalisation I}--}
Performance of the \emph{HLP} was tested in simulation scenes with the same two-row structure used in the \emph{VR} demonstrations. Here, the generalisation element involved varying the dimensions of the table and objects. Table \ref{table:controlled_experiment} summarises results of $100$ scenes from this category. The \emph{HLP} substantially outperformed the \emph{STO} by a large margin, indexed through success rates and a reduction in planning time. 

\begin{center}
\vspace{-5mm}
\begin{table}[t]
\caption{Generalisation I Simulation Scene Results.} 
\centering 
\begin{tabular}{c c c c c} 
\hline\hline 
& Success rate(\%) & Init. time(s) & Opt. time(s) & Total time(s)\\ 
\hline \hline 
HLP  & \textbf{94}  & 0.59 & 0.97 & \textbf{1.56}\\
STO  & 84 & 0.04 & 17.84 & 17.88\\
\hline\hline
\end{tabular}
\label{table:controlled_experiment}
\end{table}
\vspace{-1mm}
\end{center}

\textbf{\emph{Generalisation II}--}
Our second test involved investigating how our approach would fare when generalising to different numbers of objects, different types of objects, and different environment settings. We considered conditions with 5, 7 and 9 objects with two shape classes: cylinders and boxes. Importantly, the start and target positions were no longer aligned in this experiment. $100$ randomly sampled scenes were generated for each object-number condition. For each random scene, we selected a random\footnote{We uniformly sampled a start point along the front edge of the table, and a corresponding random end-effector orientation (in the plane) and found the initial robot configuration using inverse kinematics} initial robot configuration. 

Success rates for 100 random scenes for each of the three object categories were computed. The rates for both planners were relatively similar, (93\%, 93\% and 95\%) for \emph{HLP} and (96\%, 96\% and 98\%) for \emph{STO} for 5, 7 and 9 objects respectively. Planning time comparisons are given in Table \ref{table:generalisation2}. The time required for generating a high-level plan by \emph{HLP} (Init. Time) is fixed irrespective of the number of objects.

\begin{center}
\vspace{-2mm}
\begin{table}
\caption{Planning times (mean) for Generalisation II. The main planning time of HLP (Init. time) is almost fixed irrespective of the number of obstacles.} 
\centering 
\begin{tabular}{c c c c c} 
\hline\hline 
\multirow{2}{*}{No. of Objects} & \multicolumn{3}{c}{HLP} & STO\\
 & Init. time(s) & Opt. time(s) & Total(s) & Total(s)\\
\hline\hline
5 & 0.60 & 1.70 & 2.30 & 1.85\\
7 & 0.61 & 2.65 & 3.26 & 3.68 \\
9 & 0.63 & 5.90 & 6.53 & 4.98 \\
\hline\hline
\end{tabular}
\label{table:generalisation2}
\end{table}
\vspace{-6mm}
\end{center}

\subsubsection{Real Robot Experiments}
We used a Robotiq two-finger gripper attached to a UR5 arm mounted on an omnidirectional robot (Ridgeback). Object positions were obtained using a depth camera mounted above the table. We conducted a limited number of experiments - $4$ sample scenes per 7, 9 and 11 objects. We then ran the \emph{HLP} and the \emph{STO}, producing a total of $24$ robot runs\footnote{4 scenes, 3 number-of-object categories, and 2 methods.}. Sample results are shown in Fig. \ref{fig:real_robot} where the \emph{HLP} favoured going through the free space avoiding obstacles, while the \emph{STO} was biased to its initialised straight-line path. Other scenes are presented in the accompanying video\footnote{\url{https://youtu.be/aMIZP\_SYa0I}}.

\section{DISCUSSION}
This work speaks to the potential of human-like computing i.e. the endowment of systems with capabilities derived from modelling human perception, cognition and action. Humans are known to recognise the spatio-temporal world in a qualitative manner. Thus, instead of cloning the human behaviour from demonstrations, we used \emph{QSR} in order to segment demonstrations in the action space. We have shown that extracting human skills at such a high level helps to model a planner that can: (1) generalise to different numbers of objects without increasing the actual planning time; and (2) seamlessly connect with an arbitrary robot model.

Many cluttered environments can be clustered into regions that geometrically approximate to our definition of rows. For an arbitrary number of rows, the \emph{HLP} can be run recursively for row pairs by defining a set of sub-targets. Moreover, generalisation may be improved by augmenting training data with more generalised scenes, using more powerful classifiers, and running in a closed-loop manner. Finally, we note that further gains for this human-like approach may be made through a more detailed mechanistic understanding of the processes underlying human planning and execution. These topics will be addressed in our future work. 

\section{CONCLUSIONS}
We used \emph{VR} to collect data from human participants whilst they reached for objects on a cluttered table-top. From these data, we devised a qualitative representation of the task space to segment demonstrations into keypoints in the action space. State-action pairs were used to train decision classifiers that constituted the building blocks of the \emph{HLP} algorithm. \emph{VR}, robot simulation and real robot experiments tested against a state-of-the-art planner has shown that this \emph{HLP} is able to generate effective strategies for planning, irrespective of the number of objects in a cluttered environment. We conclude that the optimisation of robot planning has much to gain by extracting features from human action selection and execution. 

\section*{ACKNOWLEDGMENTS}
This work is funded by the EPSRC (EP/R031193/1) under their Human Like Computing initiative. The 3rd author was supported by an EPSRC studentship (1879668) and the EU Horizon 2020 research and innovation programme under grant agreement 825619 (AI4EU). The 7th, 8th and 9th authors are supported by Fellowships from the Alan Turing Institute.

\bibliographystyle{IEEEtran}
\bibliography{IEEEabrv,my_ref.bib}

\begin{thebibliography}{10}
\providecommand{\url}[1]{#1}
\csname url@samestyle\endcsname
\providecommand{\newblock}{\relax}
\providecommand{\bibinfo}[2]{#2}
\providecommand{\BIBentrySTDinterwordspacing}{\spaceskip=0pt\relax}
\providecommand{\BIBentryALTinterwordstretchfactor}{4}
\providecommand{\BIBentryALTinterwordspacing}{\spaceskip=\fontdimen2\font plus
\BIBentryALTinterwordstretchfactor\fontdimen3\font minus
  \fontdimen4\font\relax}
\providecommand{\BIBforeignlanguage}[2]{{%
\expandafter\ifx\csname l@#1\endcsname\relax
\typeout{** WARNING: IEEEtran.bst: No hyphenation pattern has been}%
\typeout{** loaded for the language `#1'. Using the pattern for}%
\typeout{** the default language instead.}%
\else
\language=\csname l@#1\endcsname
\fi
#2}}
\providecommand{\BIBdecl}{\relax}
\BIBdecl

\bibitem{latombe2012robot}
J.-C. Latombe, \emph{Robot motion planning}.\hskip 1em plus 0.5em minus
  0.4em\relax Springer Science \& Business Media, 2012, vol. 124.

\bibitem{king2015nonprehensile}
J.~E. King, J.~A. Haustein, S.~Srinivasa, and T.~Asfour, ``Nonprehensile whole
  arm rearrangement planning on physics manifolds,'' in \emph{ICRA}, 2015.

\bibitem{moll2017randomized}
M.~Moll, L.~Kavraki, J.~Rosell \emph{et~al.}, ``Randomized physics-based motion
  planning for grasping in cluttered and uncertain environments,'' \emph{IEEE
  Robotics and Automation Letters}, vol.~3, no.~2, pp. 712--719, 2017.

\bibitem{karaman2011sampling}
S.~Karaman and E.~Frazzoli, ``Sampling-based algorithms for optimal motion
  planning,'' \emph{The international journal of robotics research}, vol.~30,
  no.~7, pp. 846--894, 2011.

\bibitem{elbanhawi2014sampling}
M.~Elbanhawi and M.~Simic, ``Sampling-based robot motion planning: A review,''
  \emph{Ieee access}, vol.~2, pp. 56--77, 2014.

\bibitem{ichter2018learning}
B.~Ichter, J.~Harrison, and M.~Pavone, ``Learning sampling distributions for
  robot motion planning,'' in \emph{2018 IEEE International Conference on
  Robotics and Automation (ICRA)}.\hskip 1em plus 0.5em minus 0.4em\relax IEEE,
  2018, pp. 7087--7094.

\bibitem{fox2019multi}
R.~Fox, R.~Berenstein, I.~Stoica, and K.~Goldberg, ``Multi-task hierarchical
  imitation learning for home automation,'' in \emph{2019 IEEE 15th
  International Conference on Automation Science and Engineering (CASE)}.\hskip
  1em plus 0.5em minus 0.4em\relax IEEE, 2019, pp. 1--8.

\bibitem{zhang2018deep}
T.~Zhang, Z.~McCarthy, O.~Jow, D.~Lee, X.~Chen, K.~Goldberg, and P.~Abbeel,
  ``Deep imitation learning for complex manipulation tasks from virtual reality
  teleoperation,'' in \emph{2018 IEEE International Conference on Robotics and
  Automation (ICRA)}.\hskip 1em plus 0.5em minus 0.4em\relax IEEE, 2018, pp.
  1--8.

\bibitem{rana2018towards}
M.~Rana, M.~Mukadam, S.~R. Ahmadzadeh, S.~Chernova, and B.~Boots, ``Towards
  robust skill generalization: Unifying learning from demonstration and motion
  planning,'' in \emph{Intelligent robots and systems}, 2018.

\bibitem{tan2011computational}
H.~Tan and K.~Kawamura, ``A computational framework for integrating robotic
  exploration and human demonstration in imitation learning,'' in \emph{2011
  IEEE International Conference on Systems, Man, and Cybernetics}.\hskip 1em
  plus 0.5em minus 0.4em\relax IEEE, 2011, pp. 2501--2506.

\bibitem{ravichandar2016learning}
H.~Ravichandar and A.~Dani, ``Learning contracting nonlinear dynamics from
  human demonstration for robot motion planning,'' in \emph{ASME 2015 Dynamic
  Systems and Control Conference}.\hskip 1em plus 0.5em minus 0.4em\relax
  American Society of Mechanical Engineers Digital Collection, 2016.

\bibitem{lawitzky2012feedback}
M.~Lawitzky, J.~R. Medina, D.~Lee, and S.~Hirche, ``Feedback motion planning
  and learning from demonstration in physical robotic assistance: differences
  and synergies,'' in \emph{2012 IEEE/RSJ International Conference on
  Intelligent Robots and Systems}.\hskip 1em plus 0.5em minus 0.4em\relax IEEE,
  2012, pp. 3646--3652.

\bibitem{wallgrun2013understanding}
J.~O. Wallgr{\"u}n, F.~Dylla, A.~Klippel, and J.~Yang, ``Understanding human
  spatial conceptualizations to improve applications of qualitative spatial
  calculi,'' in \emph{27th International Workshop on Qualitative Reasoning},
  2013, p. 131.

\bibitem{chen2015survey}
J.~Chen, A.~G. Cohn, D.~Liu, S.~Wang, J.~Ouyang, and Q.~Yu, ``A survey of
  qualitative spatial representations,'' \emph{The Knowledge Engineering
  Review}, vol.~30, no.~1, pp. 106--136, 2015.

\bibitem{cohn2012thinking}
A.~G. Cohn, J.~Renz, and M.~Sridhar, ``Thinking inside the box: A comprehensive
  spatial representation for video analysis,'' in \emph{Thirteenth
  International Conference on the Principles of Knowledge Representation and
  Reasoning}, 2012.

\bibitem{mansouri2014more}
M.~Mansouri and F.~Pecora, ``More knowledge on the table: Planning with space,
  time and resources for robots,'' in \emph{2014 IEEE International Conference
  on Robotics and Automation (ICRA)}.\hskip 1em plus 0.5em minus 0.4em\relax
  IEEE, 2014, pp. 647--654.

\bibitem{westphal2011guiding}
M.~Westphal, C.~Dornhege, S.~W{\"o}lfl, M.~Gissler, and B.~Nebel, ``Guiding the
  generation of manipulation plans by qualitative spatial reasoning,''
  \emph{Spatial Cognition \& Computation}, vol.~11, no.~1, pp. 75--102, 2011.

\bibitem{csucan2011sampling}
I.~A. {\c{S}}ucan and L.~E. Kavraki, ``A sampling-based tree planner for
  systems with complex dynamics,'' \emph{IEEE Transactions on Robotics},
  vol.~28, no.~1, pp. 116--131, 2011.

\bibitem{geraerts2004comparative}
R.~Geraerts and M.~H. Overmars, ``A comparative study of probabilistic roadmap
  planners,'' in \emph{Algorithmic Foundations of Robotics V}.\hskip 1em plus
  0.5em minus 0.4em\relax Springer, 2004, pp. 43--57.

\bibitem{karaman2011anytime}
S.~Karaman, M.~R. Walter, A.~Perez, E.~Frazzoli, and S.~Teller, ``Anytime
  motion planning using the rrt,'' in \emph{2011 IEEE International Conference
  on Robotics and Automation}.\hskip 1em plus 0.5em minus 0.4em\relax IEEE,
  2011, pp. 1478--1483.

\bibitem{brookes2019studying}
J.~Brookes, M.~Warburton, M.~Alghadier, M.~Mon-Williams, and F.~Mushtaq,
  ``Studying human behavior with virtual reality: The unity experiment
  framework,'' \emph{Behavior research methods}, pp. 1--9, 2019.

\bibitem{akgun2012keyframe}
B.~Akgun, M.~Cakmak, K.~Jiang, and A.~L. Thomaz, ``Keyframe-based learning from
  demonstration,'' \emph{International Journal of Social Robotics}, vol.~4,
  no.~4, pp. 343--355, 2012.

\bibitem{bry2011rapidly}
A.~Bry and N.~Roy, ``Rapidly-exploring random belief trees for motion planning
  under uncertainty,'' in \emph{2011 IEEE international conference on robotics
  and automation}.\hskip 1em plus 0.5em minus 0.4em\relax IEEE, 2011, pp.
  723--730.

\bibitem{moldovan2018relational}
B.~Moldovan, P.~Moreno, D.~Nitti, J.~Santos-Victor, and L.~De~Raedt,
  ``Relational affordances for multiple-object manipulation,'' \emph{Autonomous
  Robots}, vol.~42, no.~1, pp. 19--44, 2018.

\bibitem{alpaydin2009introduction}
E.~Alpaydin, \emph{Introduction to machine learning}.\hskip 1em plus 0.5em
  minus 0.4em\relax MIT press, 2009.

\bibitem{langford2003reducing}
J.~Langford and B.~Zadrozny, ``Reducing t-step reinforcement learning to
  classification,'' 2003.

\bibitem{Agboh_Dogar_WAFR_2018}
W.~C. {Agboh} and M.~R. {Dogar}, ``Pushing fast and slow: Task-adaptive
  planning for non-prehensile manipulation under uncertainty,'' in
  \emph{Workshop on the Algorithmic Foundations of Robotics (WAFR)}, 2018.

\bibitem{Agboh_Dogar_Humanoids_2018}
W.~C. Agboh and M.~R. Dogar, ``Real-time online re-planning for grasping under
  clutter and uncertainty,'' in \emph{IEEE-RAS 18th International Conference on
  Humanoid Robots (Humanoids)}, 2018.

\bibitem{Agboh_Daniel_Dogar_ISRR_2019}
W.~C. Agboh, D.~Ruprecht, and M.~R. Dogar, ``Combining coarse and fine physics
  for manipulation using parallel-in-time integration,'' \emph{International
  Symposium on Robotics Research (ISRR)}, 2019.

\bibitem{mujoco}
E.~Todorov, T.~Erez, and Y.~Tassa, ``Mujoco: A physics engine for model-based
  control,'' in \emph{IROS}.\hskip 1em plus 0.5em minus 0.4em\relax IEEE, 2012.

\bibitem{dm_control}
\BIBentryALTinterwordspacing
Y.~Tassa, Y.~Doron, A.~Muldal, T.~Erez, Y.~Li, D.~de~Las~Casas, D.~Budden,
  A.~Abdolmaleki, J.~Merel, A.~Lefrancq, T.~P. Lillicrap, and M.~A. Riedmiller,
  ``Deepmind control suite,'' \emph{CoRR}, vol. abs/1801.00690, 2018. [Online].
  Available: \url{http://arxiv.org/abs/1801.00690}
\BIBentrySTDinterwordspacing

\end{thebibliography}

\end{document}